\documentclass{acm_proc_article-sp}
\usepackage{subfig}
\usepackage{hyperref}
\usepackage{tikz}
\begin{document}

\usetikzlibrary{positioning}
\usetikzlibrary{arrows}
\tikzset{>=stealth}

\title{Learning Abstract Classes using Deep Learning}

\numberofauthors{3}

\author{
  % 1st. author
  \alignauthor{}
  Sebastian Stabinger\\
  \affaddr{University of Innsbruck, Institute of Computer Science}\\
  \affaddr{Technikerstrasse 21a}\\
  \affaddr{Innsbruck, Austria}\\
  \email{Sebastian.\\Stabinger@uibk.ac.at}
  % 2nd. author
  \alignauthor{}
  Antonio Rodr\'iguez-S\'anchez\\
  \affaddr{University of Innsbruck, Institute of Computer Science}\\
  \affaddr{Technikerstrasse 21a}\\
  \affaddr{Innsbruck, Austria}\\
  \email{Antonio.Rodriguez-Sanchez@uibk.ac.at}
  % 3rd. author
  \alignauthor{}
  Justus Piater\\
  \affaddr{University of Innsbruck, Institute of Computer Science}\\
  \affaddr{Technikerstrasse 21a}\\
  \affaddr{Innsbruck, Austria}\\
  \email{Justus.Piater@uibk.ac.at}
}
\date{31 July 2015}

\maketitle
\begin{abstract}
  Humans are generally good at learning abstract concepts about
  objects and scenes (e.g.\ spatial orientation, relative sizes,
  etc.). Over the last years convolutional neural networks have
  achieved almost human performance in recognizing concrete classes
  (i.e.\ specific object categories). This paper tests the performance
  of a current CNN (GoogLeNet) on the task of differentiating between
  abstract classes which are trivially differentiable for humans. We
  trained and tested the CNN on the two abstract classes of horizontal
  and vertical orientation and determined how well the network is able
  to transfer the learned classes to other, previously unseen objects.
\end{abstract}

\category{I.2.10}{Vision and Scene Understanding}{Shape}
\category{I.5.4}{Applications}{Computer vision}
\category{I.4.8}{Scene Analysis}{Shape}
\terms{Experimentation, Performance}
\keywords{Deep Learning, Convolutional Neural Networks, Visual Cortex, Abstract Reasoning}

\section{Introduction}
Deep learning methods have gained interest from the machine learning
and computer vision research community over the last years because
these methods provide exceptional performance in classification tasks.
Especially Convolutional Neural Networks (CNNs) --- first introduced
in 1989 by LeCun et el.~\cite{lecun1989backpropagation} --- have
become popular for object classification. CNNs were more widely used
after the deep CNN from Krizhevsky et
al.~\cite{krizhevsky2012imagenet} outperformed the state of the art
methods in ILSVRC12~\cite{ILSVRC15} by a wide margin in 2012.

Convolutional neural networks consist of multiple layers of nodes,
also called neurons. One important layer type is the convolutional
layer from which the networks obtain their name. In a convolutional
layer the responses of the nodes depend on the convolution of a region
of the input image with a kernel. Additional layers introduce
non-linearities, rectification, pooling, etc. The goal of training a
CNN lies in optimizing the network weights using image-label pairs to
best reconstruct the correct label given an image. During testing the
network is confronted with novel images and expected to generate the
correct label. The network is trained by gradient descent. The
gradient is calculated by backpropagation of labeling errors. The
general idea of CNNs is to automatically learn the features needed to
distinguish classes and generate increasingly abstract features as the
information moves up the layers.

Since CNNs are very popular at the moment and perceived --- in parts
of the computer vision community --- as obtaining human like
performance we wanted to test their applicability on visual tasks
slightly outside the mainstream which are still trivially solvable by
humans. We chose to learn simple abstract classes using a standard CNN
not because we assume that they will perform better on the tasks than
other, possibly much simpler methods, but because we want to gain
insights into CNNs and how they perform on tasks which can be solved
trivially by humans. We will mainly try to give insight into the
amount of training images needed and how well the classifier
generalizes to previously unseen shapes representing the same abstract
concepts. Until now, most of the classes used for training and testing
convolutional neural networks were concrete (e.g.\ detecting classes
of objects, animal species in an image, \dots). One notable exception
is the work by G{\"u}l{\c{c}}ehre et al.\cite{gulccehre2013knowledge}
who trained a CNN to recognize whether multiple presented shapes are
the same. This is in essence a training on two abstract classes.

The problems presented by Bongard~\cite{bongard1970} inspired us to do
the research presented in this paper.
Foundalis~\cite{foundalis2006phaeaco} gives a good introduction to
these problems and presents a system intended to solve them
computationally. He used other methods than deep learning though.
Previously, Fleuret et al.\cite{fleuret2011comparing} compared human
performance to classical machine learning methods (Adaboost on
decision stumps and Support Vector Machines) on classification tasks.
The classes used were similar in spirit to Bongard problems (e.g.\
object similarity, relative position, \dots{}). Problems of similar
nature are also often given at aptitude and intelligence tests to
measure the ability for abstract, non-verbal reasoning.

We decided to use the two classes \textit{horizontal} and
\textit{vertical} for our classification experiments since they are
abstract, visually unambiguous, easy to differentiate by humans, and
easily transfer to very different shapes. The goal is to learn a
classifier that can distinguish between horizontally and vertically
oriented structures and can transfer this knowledge to previously
unseen objects and shapes. These are the first experiments exploring
the representational capabilities of current deep learning systems
regarding abstract classes.

\section{Materials and Methods}
\label{sec:materials-methods}
For this paper we used the convolutional neural network GoogLeNet as
presented by Szegedy et al.~\cite{szegedy2014going}. It won in a
number of categories in ILSVRC14. We slightly adapted the
implementation provided with the Caffe~\cite{jia2014caffe} deep
learning framework to our task (i.e.\ differentiating horizontal from
vertical shapes).

For all experiments we started with an initial learning rate of 0.01
and use ADAGRAD~\cite{duchi2011adaptive} to adapt the learning rate
over time. We trained the CNN for 1000 iterations. At this point the
loss was so small ($< 0.01$) that no further meaningful improvement
was possible. The CNN was trained 10 times on sets of different
randomly generated images to judge the mean accuracy as well as the
variance for different numbers of training images. All the graphs in
this paper show the mean accuracy as blue dots and 90\% of all
measurements fall within the shaded area (see
\autoref{fig:rect_ellipse_white} for an example). The test set
contained 250 images per class. The reported accuracy is the
proportion of correctly classified images of this test set.

\section{Experiments}
To test how well GoogLeNet can generalize abstract classes to
different shapes or different renderings we use the following
procedure: We train GoogLeNet without pre-training on a dataset
consisting of the two classes ``horizontal'' and ``vertical''. We then
test the performance of the net on a test set containing the same two
classes but represented by different shapes or rendered differently
(e.g.\ outline of the shape versus a filled representation).

We are interested whether the CNN can distinguish the two classes and
the amount of training images we need to obtain satisfactory results.

\subsection{Learning on Rectangles, Testing on Ellipses}
\label{sec:learn-rect-test}
\begin{figure}
  \centering
  \subfloat{\fbox{\includegraphics[width=0.1\textwidth]{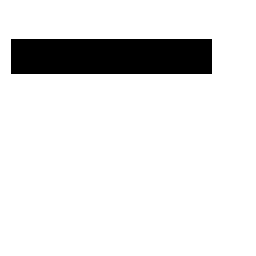}}}
  \hfil
  \subfloat{\fbox{\includegraphics[width=0.1\textwidth]{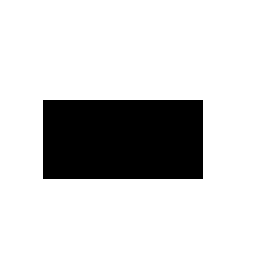}}}
  \hfil
  \subfloat{\fbox{\includegraphics[width=0.1\textwidth]{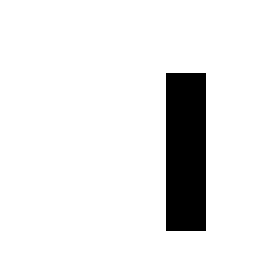}}}
  \hfil
  \subfloat{\fbox{\includegraphics[width=0.1\textwidth]{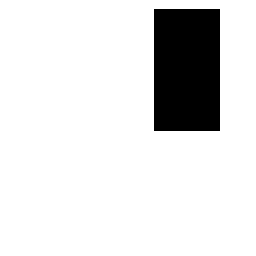}}}
  
  \subfloat{\fbox{\includegraphics[width=0.1\textwidth]{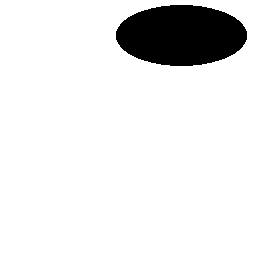}}}
  \hfil
  \subfloat{\fbox{\includegraphics[width=0.1\textwidth]{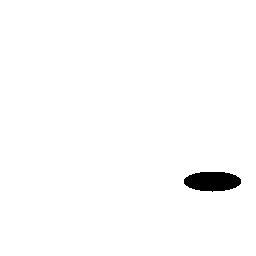}}}
  \hfil
  \subfloat{\fbox{\includegraphics[width=0.1\textwidth]{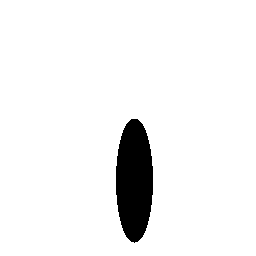}}}
  \hfil
  \subfloat{\fbox{\includegraphics[width=0.1\textwidth]{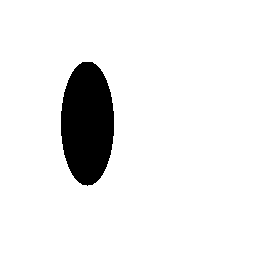}}}
  \caption{Examples of randomly generated, filled rectangles and
    ellipses (horizontal and vertical class)\label{fig:rectellipse}}
\end{figure}

We randomly generated vertically and horizontally oriented, filled
rectangles on a white background for training the network. We tested
on randomly generated, vertically or horizontally oriented, filled
ellipses (\autoref{fig:rectellipse}).

\begin{figure}[ht]
  \centering
  \includegraphics[width=0.5\textwidth]{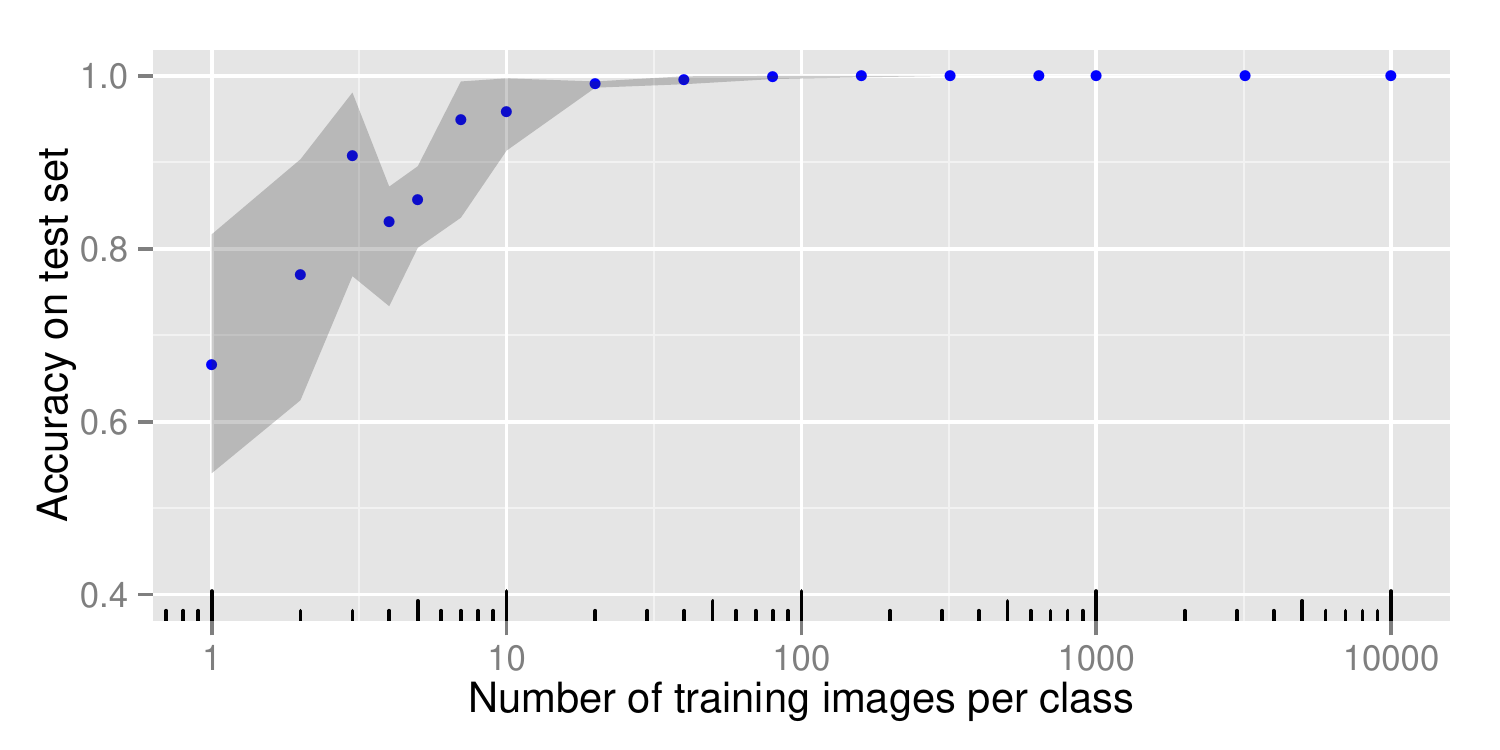}
  \caption{Learned on filled rectangles, tested on filled ellipses.
    Achieved accuracy depending on the number of training images per
    class.\label{fig:rect_ellipse_white}}
\end{figure}

\autoref{fig:rect_ellipse_white} shows the accuracy of the net after
1000 iterations in relation to the number of training images used per
class. The CNN was able to learn and generalize the two classes more
or less perfectly with about 100 training images per class. To our
surprise, even 10 images per class result in a mean accuracy of about
90\%.

As can be expected, the variance is higher for fewer training images.
One has to assume that some images are better representations of the
classes than others. Since the images are randomly generated the
quality of the whole dataset will vary more for fewer images.

\subsection{Learning on Outline, Testing on Filled}
\label{sec:outline-filled}

\begin{figure}
  \centering
  \subfloat{\fbox{\includegraphics[width=0.1\textwidth]{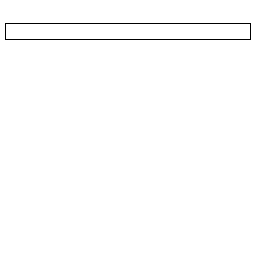}}}
  \hfil
  \subfloat{\fbox{\includegraphics[width=0.1\textwidth]{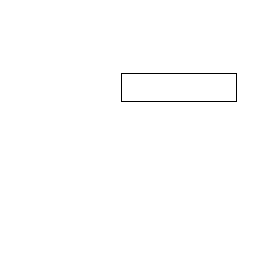}}}
  \hfil
  \subfloat{\fbox{\includegraphics[width=0.1\textwidth]{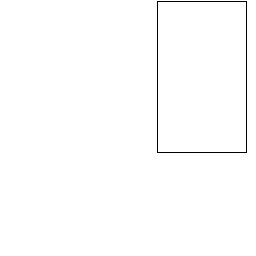}}}
  \hfil
  \subfloat{\fbox{\includegraphics[width=0.1\textwidth]{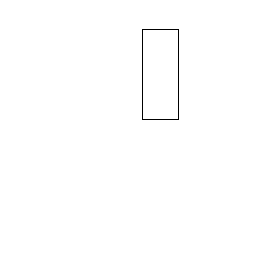}}}
  \caption{Examples of randomly generated rectangle outlines
    (horizontal and vertical class)\label{fig:outline}}
\end{figure}

To test how sensitive the network is regarding different
representations of the same shape we trained the network for the
``horizontal'' and ``vertical'' classes on outlines of rectangles
(\autoref{fig:outline}). We used the filled rectangles from
\autoref{fig:rectellipse} for testing.

\begin{figure}[ht]
  \centering
  \includegraphics[width=0.5\textwidth]{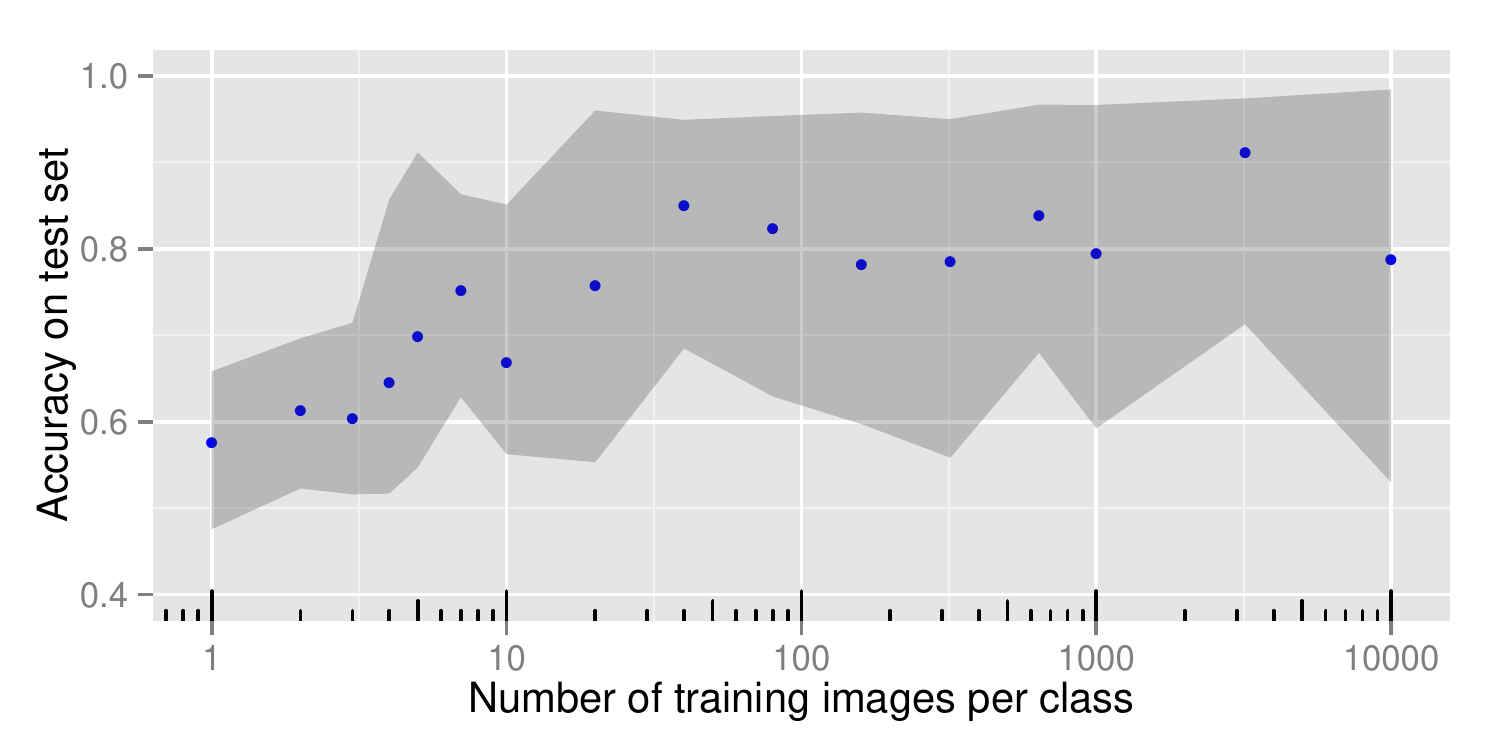}
  \caption{Learned on rectangle outlines, tested on filled rectangles.
    Achieved accuracy depending on the number of training images per
    class.\label{fig:rect_outline_rect}}
\end{figure}

\autoref{fig:rect_outline_rect} shows that the network has much bigger
problems to generalize from outlines to filled versions of the same
shape than it has at generalizing from one filled shape to another
(i.e.\ from rectangle to ellipse). In addition, the variance does not
decrease with the number of training images. This might indicate that
the network is learning features that do not capture the abstract
concept but specific information for outlines instead (i.e.\
overfitting to the training set).

\begin{figure}[ht]
  \centering
  \includegraphics[width=0.5\textwidth]{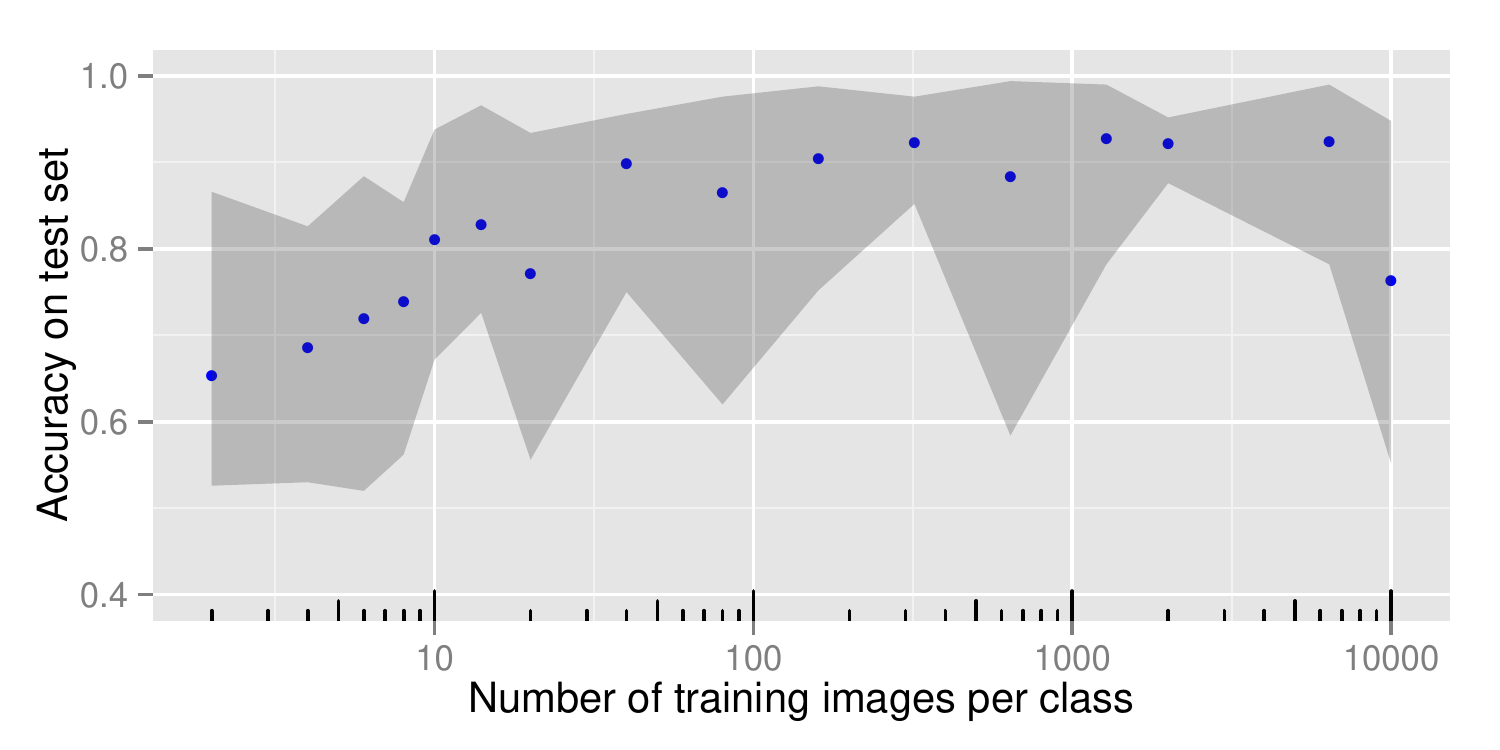}
  \caption{Learned on rectangle and ellipse outlines, tested on filled
    rectangles. Achieved accuracy depending on the number of training
    images per class.\label{fig:rectellipse_outline_rect}}
\end{figure}

If a CNN is able to learn abstract concepts one can reason that adding
another shape outline to the training set will improve the accuracy.
By this we force the net to learn a more abstract concept. To test
this hypothesis we added horizontally and vertically oriented ellipse
outlines to the training images and again tested the accuracy on the
filled rectangles. As can we can see in
\autoref{fig:rectellipse_outline_rect}, the addition of ellipse
outlines improved the performance on the filled rectangle test set
slightly.

\subsection{Random Shapes}
\label{sec:random-shapes}

\begin{figure}
  \centering
  \subfloat{\fbox{\includegraphics[width=0.1\textwidth]{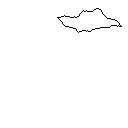}}}
  \hfil
  \subfloat{\fbox{\includegraphics[width=0.1\textwidth]{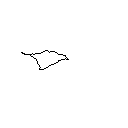}}}
  \hfil
  \subfloat{\fbox{\includegraphics[width=0.1\textwidth]{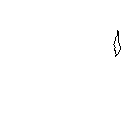}}}
  \hfil
  \subfloat{\fbox{\includegraphics[width=0.1\textwidth]{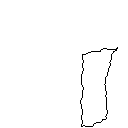}}}
  \caption{Examples of randomly generated shapes (horizontal and
    vertical class)\label{fig:randshape}}
\end{figure}

We performed the last set of experiments on random shapes
(\autoref{fig:randshape}) which we created with an adapted version of
the SVRT framework presented by Fleuret et
al.~\cite{fleuret2011comparing}.

\begin{figure}[ht]
  \centering
  \includegraphics[width=0.5\textwidth]{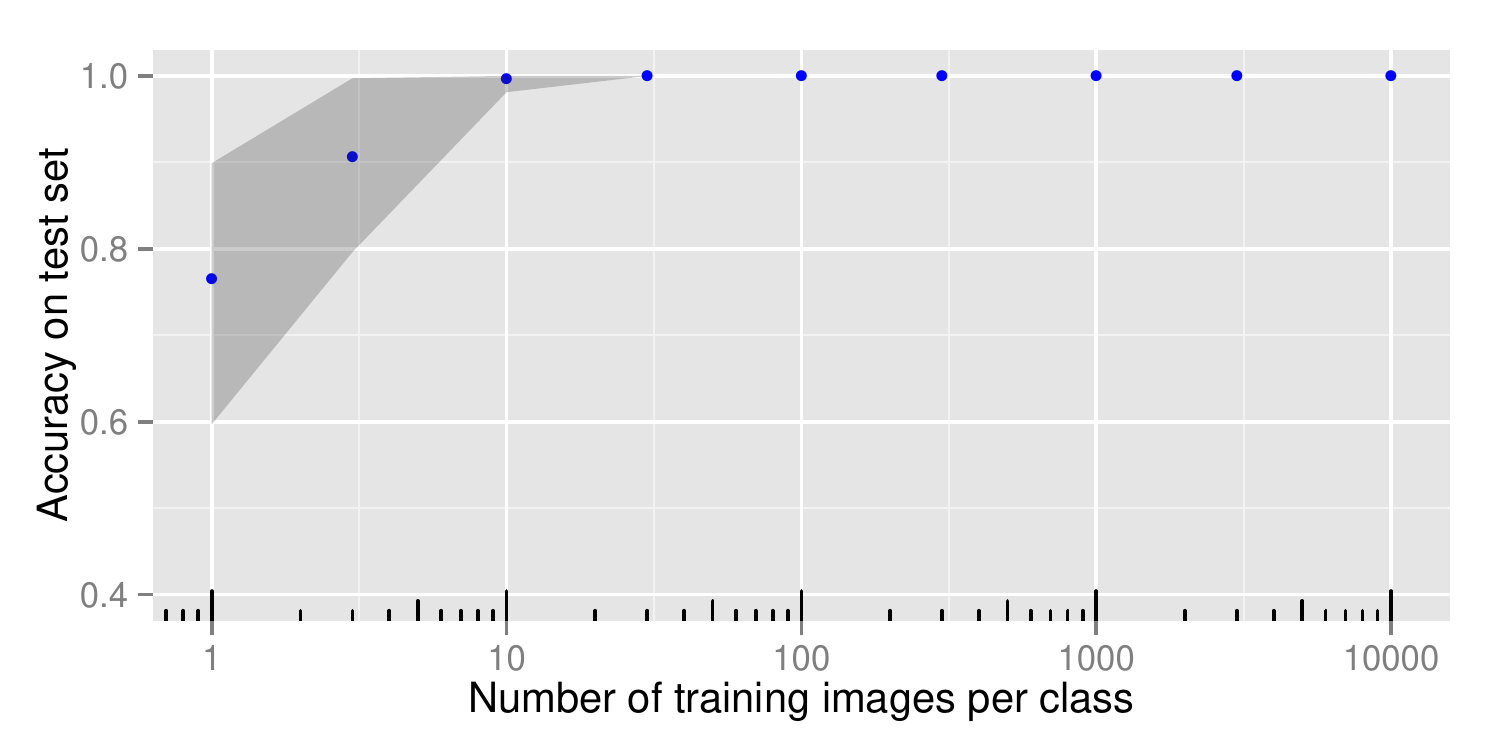}
  \caption{Learned on random outlines, tested on random outlines.
    Achieved accuracy depending on the number of training images per
    class.\label{fig:rnd_outline_rnd_outline} }
\end{figure}

The network is able to learn the two abstract classes with these
highly varied shapes (\autoref{fig:rnd_outline_rnd_outline}) and is
also able to transfer the knowledge from outlines to filled shapes
(\autoref{fig:rnd_outline_rnd_filled}).

\begin{figure}
  \centering
  \subfloat{\fbox{\includegraphics[width=0.1\textwidth]{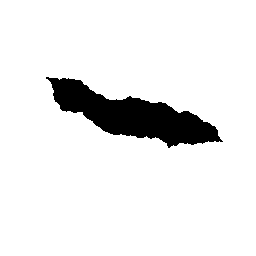}}}
  \hfil
  \subfloat{\fbox{\includegraphics[width=0.1\textwidth]{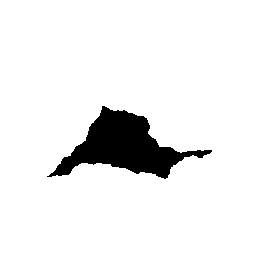}}}
  \hfil
  \subfloat{\fbox{\includegraphics[width=0.1\textwidth]{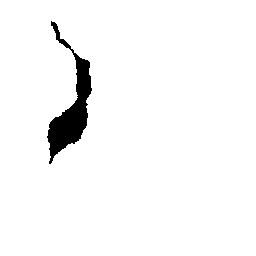}}}
  \hfil
  \subfloat{\fbox{\includegraphics[width=0.1\textwidth]{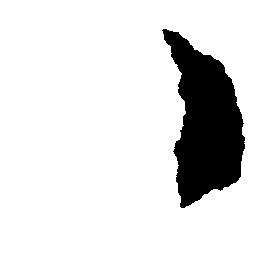}}}
  \caption{Examples of randomly generated, filled shapes (horizontal
    and vertical class)\label{fig:randshapefilled}}
\end{figure}

\begin{figure}[ht]
  \centering
  \includegraphics[width=0.5\textwidth]{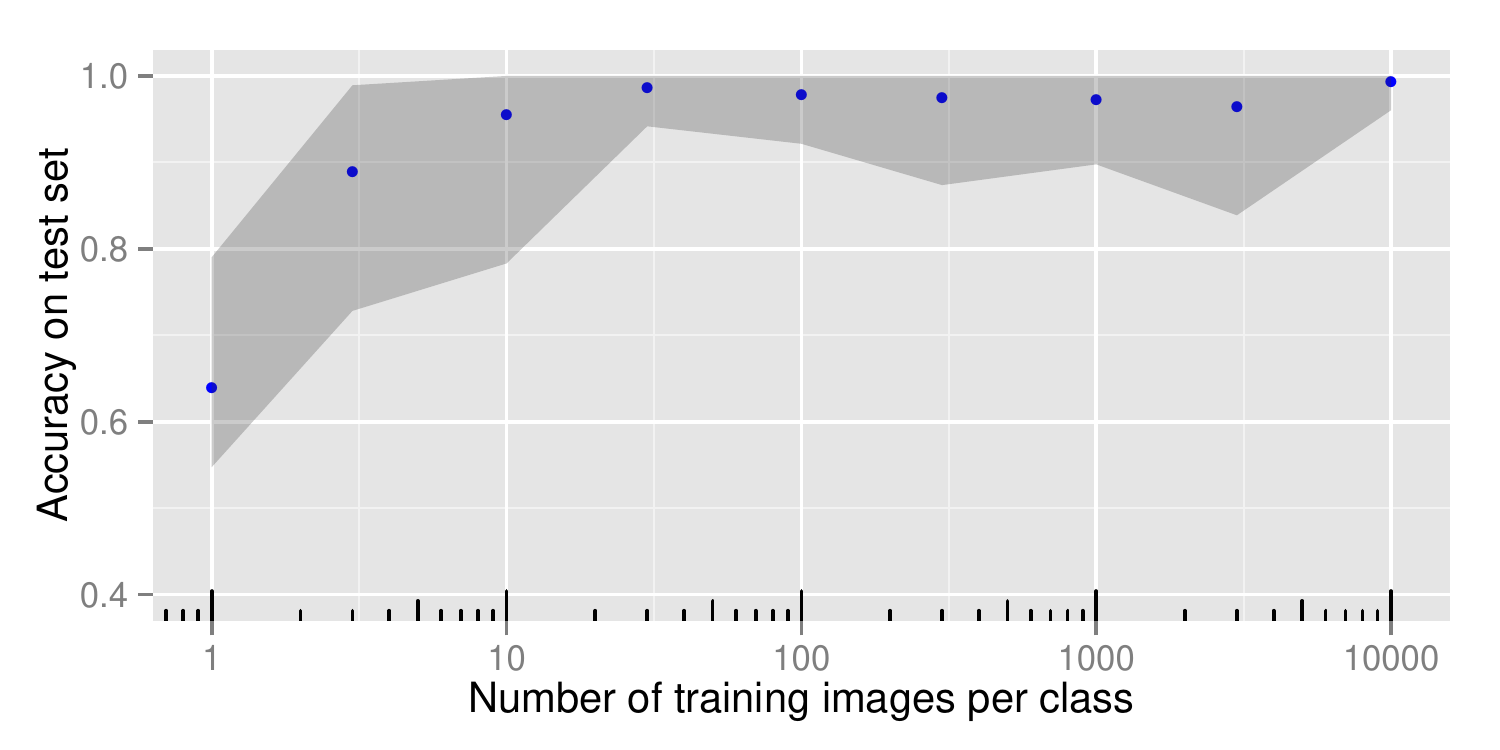}
  \caption{Learned on random outlines, tested on random, filled
    shapes. Achieved accuracy depending on the number of training
    images per class.\label{fig:rnd_outline_rnd_filled} }
\end{figure}

\begin{figure}[ht]
  \centering
  \includegraphics[width=0.5\textwidth]{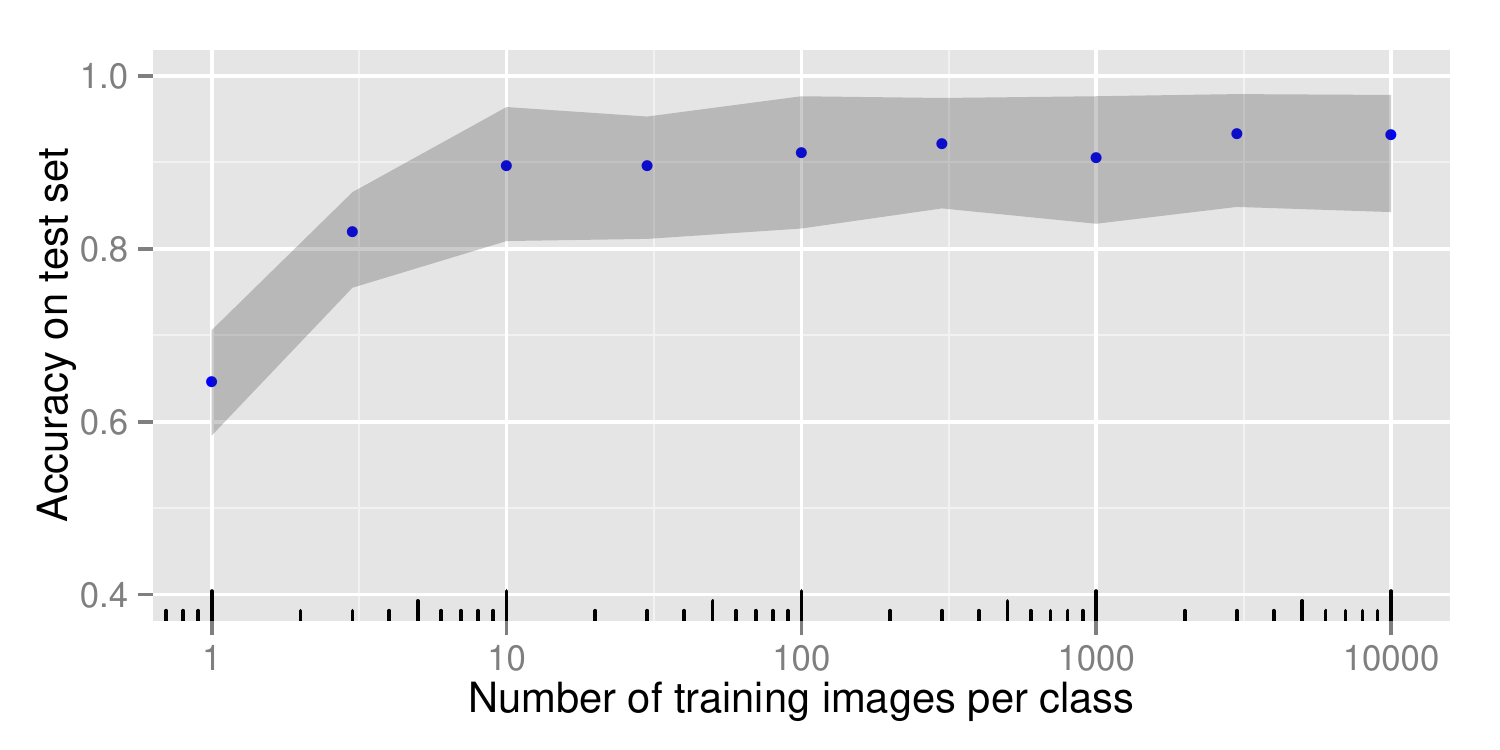}
  \caption{Learned on random outlines, tested on filled rectangles.
    Achieved accuracy depending on the number of training images per
    class.\label{fig:rnd_outline_rect_filled} }
\end{figure}

\autoref{fig:rnd_outline_rect_filled} shows that the network trained
on random shape outlines even performs better at detecting the
orientation of filled rectangles than the network trained on similar
data sets (\autoref{fig:rect_outline_rect} and
\autoref{fig:rectellipse_outline_rect})

\begin{figure}
  \centering
  \subfloat{\fbox{\includegraphics[width=0.1\textwidth]{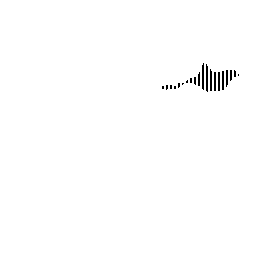}}}
  \hfil
  \subfloat{\fbox{\includegraphics[width=0.1\textwidth]{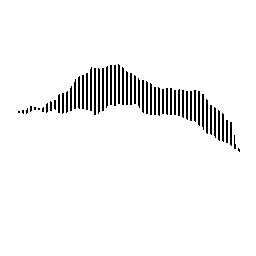}}}
  \hfil
  \subfloat{\fbox{\includegraphics[width=0.1\textwidth]{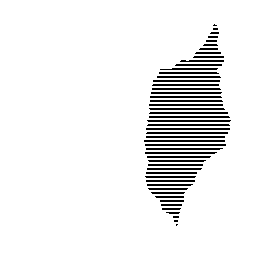}}}
  \hfil
  \subfloat{\fbox{\includegraphics[width=0.1\textwidth]{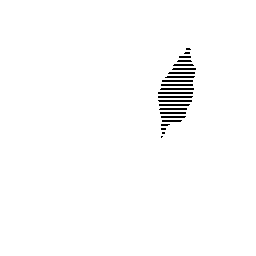}}}
  \caption{Examples of randomly generated, textured shapes (horizontal
    and vertical class)\label{fig:textured}}
\end{figure}

\begin{figure}[ht]
  \centering
  \includegraphics[width=0.5\textwidth]{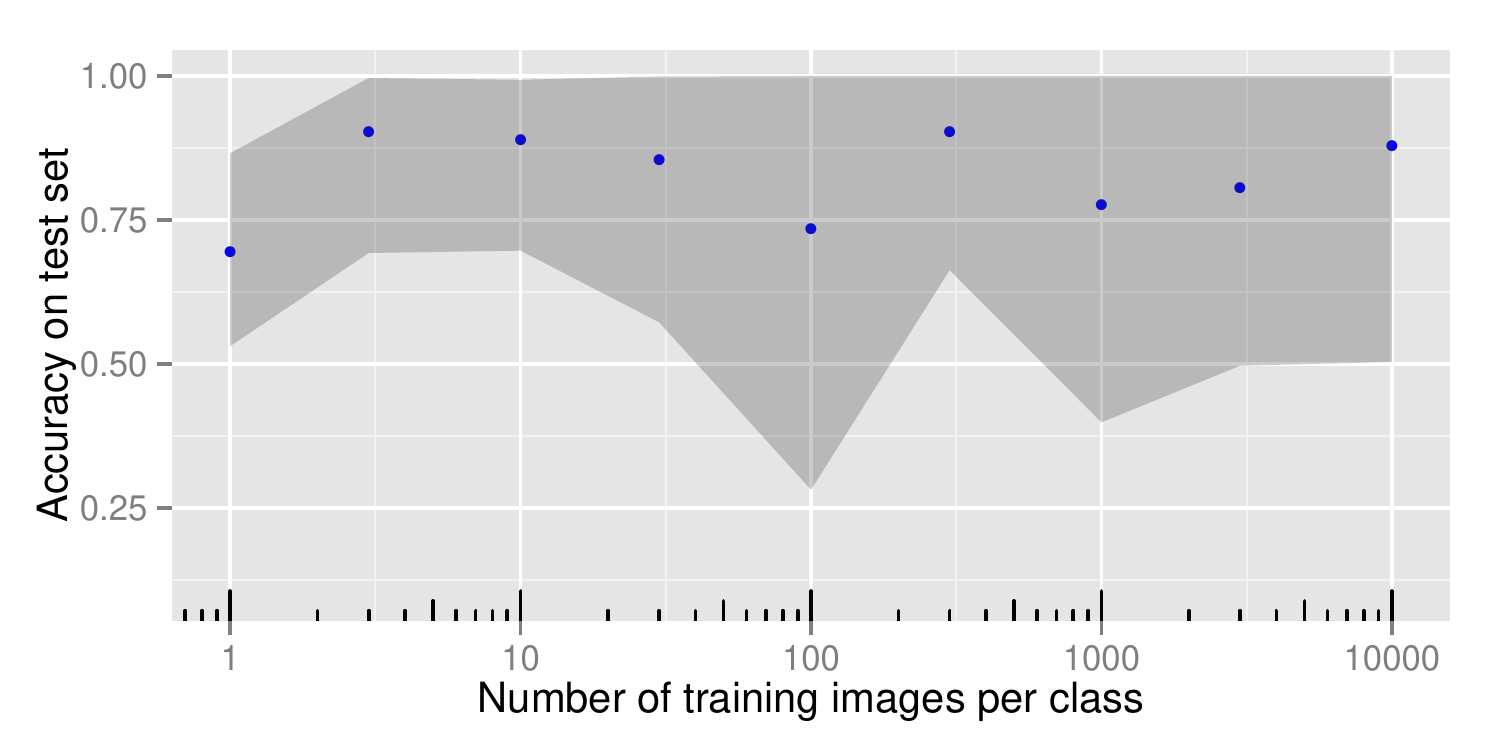}
  \caption{Learned on random outlines, tested on texture filled random
    shapes. Achieved accuracy depending on the number of training
    images per class.\label{fig:rnd_rnd_texture}}
\end{figure}

As a final experiment we looked at how well the random shapes
generalize to other, textured random shapes. We used the likely most
difficult test set where the texture orientation was orthogonal to the
orientation of the shape. \autoref{fig:textured} shows examples of
this class. The results (\autoref{fig:rnd_rnd_texture}) indicate that
more training examples lead to extreme variance in the results. The
performance of the network varies from perfect accuracy to pure
guessing. In addition, nothing during the training phase indicates how
well the network will perform on the test set (e.g.\ lower loss during
training is uncorrelated to the performance on the test set).

\section{Conclusion}
\label{sec:conclusion}
We showed that a state of the art convolutional neural network is able
to learn abstract classes and transfer that information to other,
previously unseen, shapes. But it is also apparent that the current
networks are sensitive to the used training and testing data regarding
how well the transfer of knowledge works.

Probably the best example is the training on filled rectangles and
testing on filled ellipses in comparison to training on rectangle
outlines and testing on filled rectangles. It was unclear before doing
experiments that GoogLeNet will perform much better on the first task
than the second one. Humans are in general much less affected by such
representational differences and perform well on highly variable data
sets as can be seen in problem sets used for measuring non-verbal
abstract reasoning or the Bongard problems.

Of course humans as well as animals already have pre-training before
encountering such tasks. Pre-training in the form of previously
learned concepts as well as the optimization that occurred during
evolution and manifests itself in the organization of the brain. A CNN
is missing this information. We think therefore that pre-training of
networks on the right data set will be paramount to increase the
performance on more abstract tasks. The results of G{\"u}l{\c{c}}ehre
et al.\cite{gulccehre2013knowledge} also point into this direction.

\bibliographystyle{abbrv}
\bibliography{sigproc}

\section{Appendix --- GoogLeNet}
\label{sec:appendix}
In this appendix we will give a brief introduction to GoogLeNet, the
convolutional neural network used for the experiments in this paper.

\subsection{Inception Module}
\label{sec:inception-module}
\begin{figure}[ht]
  \centering
  \begin{tikzpicture}[node distance=0.8cm,>=stealth, scale=0.6, every node/.style={transform shape}]
    \tikzstyle{state_blue}=[thick,draw=blue!75,fill=blue!20,text width=2cm, align=center]
    \tikzstyle{state_yellow}=[thick,draw=yellow!75,fill=yellow!20,text width=2cm, align=center]
    \tikzstyle{state_red}=[thick,draw=red!75,fill=red!20,text width=2cm, align=center]
    \tikzstyle{state_green}=[thick,draw=green!75,fill=green!20,text width=2cm, align=center]
    \tikzstyle{edge}=[->,draw=black,line width=0.2mm]

    \node[state_green] (filterconcat) at (-0.4,2) {Filter concatenation};
    \node (out) [above of=filterconcat] {};
    
    \node[state_blue] (1x1) at (-5,0) {$1 \times 1$ convolutions};
    \node[state_blue] (3x3) [right=of 1x1] {$3 \times 3$ convolutions};
    \node[state_blue] (5x5) [right=of 3x3] {$5 \times 5$ convolutions};
    \node[state_yellow] (1x1cpool) [right=of 5x5] {$1 \times 1$ convolutions};

    \node[state_yellow] (1x1c3) [below=of 3x3] {$1 \times 1$ convolutions};
    \node[state_yellow] (1x1c5) [below=of 5x5] {$1 \times 1$ convolutions};
    \node[state_red] (3x3pool) [below=of 1x1cpool] {$3 \times 3$ max pooling};

    \node[state_green] (previous) at (-0.4,-3.6) {Previous layer};
    \node (in) [below of=previous] {};

    \draw[edge][out=90, in=-90] (previous) to (1x1c3);
    \draw[edge][out=90, in=-90] (previous) to (1x1c5);
    \draw[edge][out=0, in=-90] (previous) to (3x3pool);
    \draw[edge][out=180, in=-90] (previous) to (1x1);

    \draw[edge] (1x1c3) to (3x3);
    \draw[edge] (1x1c5) to (5x5);
    \draw[edge] (3x3pool) to (1x1cpool);

    \draw[edge][out=90, in=180] (1x1) to (filterconcat);
    \draw[edge][out=90, in=-90] (3x3) to (filterconcat);
    \draw[edge][out=90, in=-90] (5x5) to (filterconcat);
    \draw[edge][out=90, in=0] (1x1cpool) to (filterconcat);

    \draw[edge] (in) to (previous);
    \draw[edge] (filterconcat) to (out);
  \end{tikzpicture}
  \caption{Overview of an inception module. Adapted from Szegedy et
    al.~\cite{szegedy2014going}\label{fig:inception}. The $1 \times 1$
    convolutions for dimensionality reduction are shown in yellow.}
\end{figure}
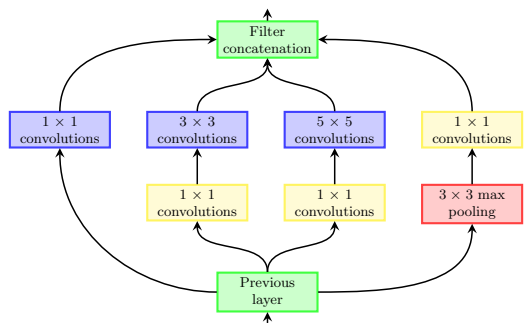

An inception module is a small two layer network which is repeated
many times. It is used as the main building block of GoogLeNet.
\autoref{fig:inception} shows a graphical representation of an
inception module.

An inception module computes convolutions with different receptive
field sizes ($1 \times 1$, $3 \times 3$, $5 \times 5$) and
$3 \times 3$ max pooling in parallel and concatenates all the
responses to produce the output to the next layer. Since this would
lead to excessive amounts of parameters, $1 \times 1$ convolutions are
used for dimensionality reduction.

\subsection{GoogLeNet}
\label{sec:googlenet}
GoogLeNet consist of a stack of nine inception modules. There are
three points at which softmax is being used to calculate the loss of
the network. One at the end of all nine inception modules and two
after the third and sixth inception module. The reasoning behind using
middle layers to calculate an error function is to promote better
discrimination in lower layers and to calculate a better gradient
signal. Both is needed since we are dealing with a very deep network
with 27 layers. We refer to Figure 3 in the paper by Szegedy et
al.~\cite{szegedy2014going} for a more detailed description of the
layer structure of GoogLeNet.

\subsection{Specifics of the Implementation}
\label{sec:spec-used-impl}
The implementation in the Caffee framework differs slightly from the
network presented by Szegedy et al.~\cite{szegedy2014going}. 

Stochastic gradient descent with momentum is used to update the
weights and the Xavier algorithm as presented by Glorot et
al.~\cite{glorot2010understanding} is used for initializing the
weights.

\balancecolumns{}
\end{document}